# Cost Sensitive Reachability Heuristics for Handling State Uncertainty

**Daniel Bryce & Subbarao Kambhampati**
Department of Computer Science and Engineering
Arizona State University, Brickyard Suite 501
699 South Mill Avenue, Tempe, AZ 85281
{dan.bryce, rao}asu.edu

## Abstract

While POMDPs provide a general platform for non-deterministic conditional planning under a variety of quality metrics they have limited scalability. On the other hand, non-deterministic conditional planners scale very well, but many lack the ability to optimize plan quality metrics. We present a novel generalization of planning graph based heuristics that helps conditional planners both scale and generate high quality plans when using actions with non-uniform costs. We make empirical comparisons with two state of the art planners to show the benefit of our techniques.

## 1 Introduction

When agents have uncertainty about their state, they need to formulate conditional plans, which attempt to resolve state-uncertainty with sensing actions. This problem has received attention in both the uncertainty in AI (UAI) and automated planning communities. From the UAI perspective, finding such conditional plans is a special case of finding policies for Markov Decision Processes (MDPs) in the fully observable case, and Partially Observable MDPs (POMDPs) in the partially observable case. The latter is of more practical use, although much harder computationally [Madani *et al.*, 1999; Littman *et al.*, 1998]. The emphasis in this community has been on finding optimal policies under fairly general conditions. However the scalability of the approaches has been very limited. In the planning community, conditional planning has been modelled as search in the space of uniform probability belief states (i.e. every belief state is a set of equally possible states and the space is finite). Several planners have been developed–eg. MBP [Bertoli *et al.*, 2001], and PKSPlan [Petrick and Bacchus, 2002] – which model conditional plan construction as an and/or search. These approaches are more scalable partly because their complexity is *only* 2-EXP-complete [Rintanen, 2004], as against POMDPs which are in general undecidable [Madani *et al.*, 1999]. However planning approaches are often insensitive to the cost/quality information. Indeed, in the presence of actions with differing costs, planners such as MBP can generate plans of arbitrarily low quality, attempt to insert sensing actions without taking their cost into consideration.

We focus our attention on finding strong plans (i.e. plans the succeed with probability 1) given an uncertain initial state (with uniform probability over possible states). Sensing actions give partial observations, causative actions have deterministic conditional effects, all actions have associated costs, and the model uses a factored representation.

In this paper, we describe a way of extending the state of the art non-deterministic conditional planners to make them more sensitive to cost/quality information. Our idea is to adapt the type of cost-sensitive reachability heuristics that have proven to be useful in classical and temporal planning [Do and Kambhampati, 2003]. Straightforward adaptation unfortunately proves to be infeasible. This is because, in the presence of state uncertainty, we will be forced to generate multiple planning graphs (one for each possible state) and reason about reachability across all those graphs [Bryce and Kambhampati, 2004]. This can get prohibitively expensive–especially for forward search where we need to do this analysis at each search node.

The main contribution of this paper is a way to solve this dilemma. In particular, we propose a novel way of generating reachability information with respect to belief states without computing multiple graphs. Our approach, called the labelled uncertainty graph ($LUG$), symbolically represents multiple planning graphs, one for each state in our belief, within a single planning graph. Loosely speaking, this single graph unions the support information present in explicit multiple graphs and pushes the disjunction, describing sets of possible worlds (states in a belief), into "labels" ($\ell$). The planning graph is built using labels, for sets of worlds, to annotate the vertices (literals and actions). A label on a vertex signifies the states of our belief that can reach the vertex.

To take cost into account, we describe a method for prop-agating cost information over the $LUG$ in an extension called the $CLUG$. The (previously mentioned) labels tell us when graph vertices (e.g. literals) are reachable, but they do not indicate the associated reachability cost. We could track a single cost for the entire set of worlds represented by a label, but this would lose information about differing costs for subsets of the worlds. Tracking a cost for each subset of worlds is also problematic because they are an exponential in the number of worlds. Even tracking the cost of individual worlds can also be costly because their number is exponential in the number of fluents (state variables). Instead we track cost over a fixed partition of world sets. The size of the partition (number of costs tracked) is bounded by the number of planning graph levels. Each disjoint set is the worlds in which a literal or action is newly reached at a level. The $CLUG$ is used as the basis for doing reachability analysis. In particular, we extract relaxed plans from it (as described in [Bryce et al., 2004]), using the cost information to select low cost relaxed plans. Our results show that cost-sensitive heuristics improve plan quality and scalability.[1]

We proceed by describing our representation and our planner, called $POND$. We then introduce our planning graph generalizations called the $LUG$ and the $CLUG$, and describe the relaxed plan extraction procedure. We present an empirical study of the techniques within our planner and compare with two state of the art conditional planners MBP [Bertoli et al., 2001] and GPT [Bonet and Geffner, 2000]. We end by providing a comparison to related work, a conclusion, and directions for future work, with emphasis on non-uniform uncertainty.

## 2 Representation & Search

The planning formulation in our planner $POND$ uses progression search to find strong plans, under the assumption of partial observability. A strong plan guarantees that after a finite number actions executed from any of the many possible initial states, all resulting states will satisfy the goals. We represent strong plans as directed acyclic graphs (where a node with out degree greater than one is a sensory action). We assume that every plan path is equally likely so our plan quality metric is the mean of the path costs. The cost of a plan path the sum of the costs of its edges (which correspond to outcomes of actions).

We will use the following as a motivating, as well as, running example to illustrate our techniques:

**Example 1.** *A patient goes to the doctor complaining of feeling unrested ($\neg r$), but he is unsure if he is actually sick ($s \vee \neg s$). The doctor has two treatment plans: 1) give the patient drug B to cure the sickness if he is sick, and have*

[1] A solution for a larger test instance contained nearly 200 belief states among 13 plan paths, of lengths between 18 and 30 actions.

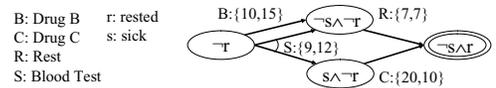

Figure 1: *The example's AO* graph with two cost models.*

*him rest, R, for a week to become rested, or 2) do a blood test, S, to determine if he is sick; if so he takes drug C with no need to rest, otherwise rests for a week. Both treatments will ensure that he is not sick and rested ($\neg s \wedge r$).*

The patient may have one of two insurance providers (cost models). We show a transition diagram (Figure 1) with annotations on edges for the two cost models. The optimal plan for the first model is the first plan, at cost 10+7 = 17, compared to the second at cost ((9+7)+(9+20))/2 = 22.5. The optimal plan for the second model is the second plan with cost ((12+7)+(12+10))/2 = 20.5, because the first has cost 15+7 = 22.

$POND$ searches in the space of belief states, a technique first described by Bonet and Geffner [2000]. The planning problem $P$ is defined as the tuple $\langle D, BS_I, BS_G \rangle$, where $D$ is a domain, $BS_I$ is the initial belief state, and $BS_G$ is the goal belief state. The domain $D$ is a tuple $\langle F, A \rangle$, where $F$ is a set of all fluents and $A$ is a set of actions.

**Belief State Representation:** A state $S$ is a complete interpretation over fluents. A belief state $BS$ is a set of states, symbolically represented as a propositional formula over $F$, and is also referred to as a set of possible worlds. A state $S$ is in the set of states represented by a belief state $BS$ if $S$ is a model of $BS$ ($S \in \mathcal{M}(BS)$). In this work we assume the goal belief state is a conjunctive formula to simplify the later presentation.

**Action Representation:** We represent actions as having strictly causative or observational effects, respectively termed as causative or sensory actions. An action $a$ consists of an execution precondition $\rho^e(a)$, a set of effects $\Phi(a)$, and a cost $c(a)$. The execution precondition, $\rho^e(a)$, is a conjunctive formula that must hold to execute the action. Causative actions have a set of deterministic conditional effects $\Phi(a) = \{\varphi^0(a), ..., \varphi^m(a)\}$ where each conditional effect $\varphi^j(a)$ is of the form $\rho^j(a) \implies \varepsilon^j(a)$, and the antecedent and consequent are conjunctions. Sensory actions have a set $\Phi(a) = \{o^0(a), ..., o^n(a)\}$ of observational effect formulas. Each observational effect formula, $o_i$, defines an outcome of the sensor.

The actions in our example are:
$B : \langle \rho^e(B) = \top, \Phi(B) = \{s \implies \neg s\}, c(B) = \{10, 15\}\rangle$
$C : \langle \rho^e(C) = s, \Phi(C) = \{\top \implies \neg s \wedge r\}, c(C) = \{20, 10\}\rangle$
$R : \langle \rho^e(R) = \neg s, \Phi(R) = \{\top \implies r\}, c(R) = \{7, 7\}\rangle$
$S : \langle \rho^e(S) = \top, \Phi(S) : \{s, \neg s\}, c(S) = \{9, 12\}\rangle$

We list two numbers in the cost of each action because our example uses the first number for cost model one, and the second for cost model two.

**POND Search:** We use top down AO* search [Nilsson, 1980], in the $POND$ planner to generate conditional plans. In the search graph, the nodes are belief states and the hyper-edges are action. We need AO* because using a sensing action in essence partitions the current belief state. We use a hyper-edge to represent the collection of outcomes of an action. Sensory actions have several outcomes, all if any of which must be included in a solution.

The AO* search consists of two repeated steps, expand the current partial solution, and then revise the current partial solution. Search ends when every leaf node of the current solution is a belief state that satisfies the goal belief and no better solution exists (given our heuristic function). Expansion involves following the current solution to an unexpanded leaf node and generating its children. Revision is essentially a dynamic programming update at each node in the current solution that selects a best hyper-edge (action). The update assigns the action with minimum cost to start the best solution rooted at the given node. The cost of a node is the cost of its best action plus the average cost of its children (the nodes connected through the hyper-edge). When expanding a leaf node, the children of all applied actions are given a heuristic value to indicate their estimated cost.

## 3 Labelled Uncertainty Graph ($LUG$)

To guide search, we use a relaxation of conditional planning to obtain a lookahead estimate of the conditional plan's suffix, rooted at each search node. The relaxation measures the cost to support the goal when ignoring mutexes between actions, and ignoring sensory actions. We reason about the cost of sensing in a local manner through the search itself,[2] but do not reason about sensing in a lookahead fashion. Our heuristic reasons about the conformant transition cost between two sets of states, a belief state cost measure [Bryce and Kambhampati, 2004]. We review our previous work that uses multiple planning graphs to calculate belief state distances, and then discuss our generalization, called the $LUG$ which performs the same task at a lower cost.

Classical planning graph based relaxed plans tend not to capture information needed to make belief state to belief state distance measures because they assume perfect state information. In [Bryce and Kambhampati, 2004] we studied the use of classical planning graphs for belief state distance measures, but found that using multiple planning graphs is more effective for estimating belief state distances. The approach constructs several classical planning graphs, each with respect to a state in our current belief state. Then a classical relaxed plan is extracted from each graph. We transform the resulting set of relaxed plans into a unioned relaxed plan, where each layer is the union over the vertices in the same level of the individual relaxed plans. The number of action vertices in the unioned relaxed plan is used as the heuristic estimate. The heuristic measures both the positive interaction and independence in action sequences that are needed to individually transition each state in our belief state to a state in the goal belief state.

The obvious downfall of the multiple graph approach is that the number of planning graphs and relaxed plans is exponential in the size of belief states. Among the multiple planning graphs there are quite a bit of repeated structure, and computing a heuristic on each can take a lot of time. With the $LUG$, our intent is two fold, (i) we would like to obtain the same heuristic as with multiple graphs, but lower the representation and heuristic extraction overhead, and (ii) we also wish to extend the relaxed plan heuristic measure to reflect non uniform action costs.

### 3.1 $LUG$ & $CLUG$

We present the $LUG$ and its extension to handle costs, the $CLUG$. The $LUG$ is a single planning graph that uses an annotation on vertices (actions and literals) to reflect assumptions about how a vertex is reached. Specifically we use a label, $(\ell_k(\cdot))$, to denote the models of our current (source) belief $BS_s$ that reach the vertex in level $k$. In the $CLUG$ we additionally use a cost vector ($c_k(\cdot)$) to estimate of the cost of reaching the vertex from different models of the source belief. These annotations help us implicitly represent the vertices common to several of the multiple planning graphs in a single planning graph. Figure 2 illustrates the $CLUG$ built for the initial belief in our example. The initial layer literal labels are used to label the actions and effects they support, which in turn label the literals they support. Label propagation is based on the intuition that (i) actions and effects are applicable in the possible worlds in which their conditions are reachable and (ii) a literal is reachable in all possible worlds where it is affected.

**Definition 1** ($LUG$). *A $LUG$ is a levelled graph, where a level $k$ contains three layers, the literal $\mathcal{L}_k$, action $\mathcal{A}_k$, and effect $\mathcal{E}_k$ layers. The $LUG$ is constructed with respect to the actions in $A$ and a source belief state $BS_s$. Each $LUG$ vertex $v_k(\cdot)$ in level $k$ is a pair $\langle \cdot, \ell_k(\cdot) \rangle$, where the "·" is an action $a$, effect $\varphi^j(a)$, or literal $l$, and $\ell_k(\cdot)$ is its label.*

**Definition 2** ($CLUG$). *A $CLUG$ extends a $LUG$ by associating a triple $\langle \cdot, \ell_k(\cdot), c_k(\cdot) \rangle$ with each vertex, where $c_k(\cdot)$ is a cost vector.*

**Definition 3** (Label). *A label $\ell_k(\cdot)$ is a propositional formula that describes a set of possible worlds. Every model of a label is also a model of the source belief, implying $\ell_k(\cdot) \models BS_s$. For any model $S_s \in \mathcal{M}(BS_s)$ if $S_s \in \mathcal{M}(\ell_k(\cdot))$, then the classical relaxed planning graph built from $S_s$ contains "·" as a vertex in level $k$.*

---

[2]That is, we can reason about the cost of applying a sensing action at the current search node by adding the cost of the action to the average cost of its children (whose costs are determined by the heuristic).

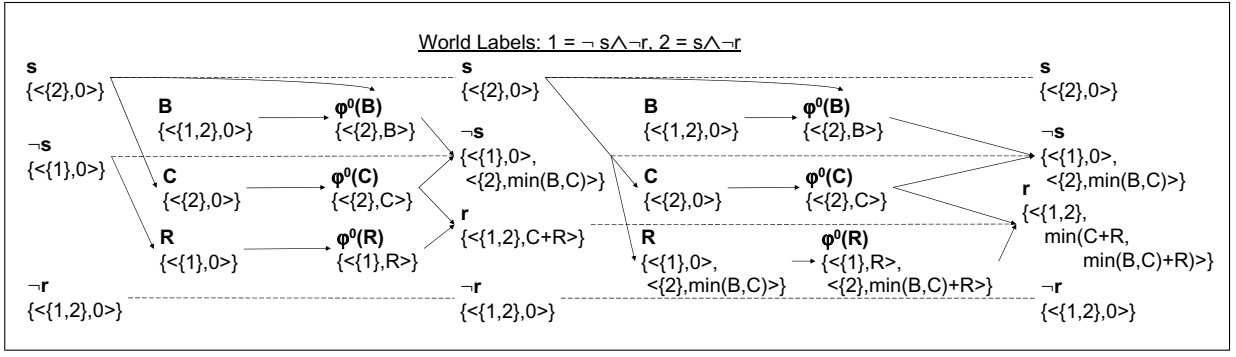

Figure 2: *A LUG for our example problem. Each literal, action, and effect has its cost vector listed.*

**Definition 4** (Extended Label). *An extended label $\ell_k^*(f)$ for a propositional formula $f$ is defined as the formula that results from substituting the label $\ell_k(l)$ of each literal $l$ for the literal in $f$. An extended label is defined:*

$$\ell_k^*(f \wedge f') = \ell_k^*(f) \wedge \ell_k^*(f'),$$
$$\ell_k^*(f \vee f') = \ell_k^*(f) \vee \ell_k^*(f'),$$
$$\ell_k^*(\neg(f \wedge f')) = \ell_k^*(\neg f \vee \neg f'),$$
$$\ell_k^*(\neg(f \vee f')) = \ell_k^*(\neg f \wedge \neg f'),$$
$$\ell_k^*(\top) = BS_s, \ell_k^*(\bot) = \bot, \ell_k^*(l) = \ell_k(l)$$

**Labels and Reachability:** *A literal $l$ is (optimistically) reachable from a set of states, described by $BS_s$, after $k$ steps, if $BS_s \models \ell_k(l)$. A propositional formula $f$ is reachable from $BS_s$ after $k$ steps if $BS_s \models \ell_k^*(f)$.*

**Definition 5** (Cost Vectors). *A cost vector $c_k(\cdot)$ is a set of pairs $\langle f^i(\cdot), c^i(\cdot) \rangle$, where $f^i(\cdot)$ is a propositional formula over $F$ and $c^i(\cdot)$ is a rational number. Every $c^i(\cdot)$ is an estimate of the cost of reaching the vertex from all models $S_s \in \mathcal{M}(f^i(\cdot))$.*

Cost propagation on planning graphs, similar to that used in the Sapa planner [Do and Kambhampati, 2003], computes the estimated cost of reaching literals at time points. Since we track whether a literal is reached in more than one possible world, it is possible that the cost of reaching a literal is different for every subset of these worlds. Instead of tracking costs for an exponential number of subsets, or even each individual world, we partition the models of $BS_s$ into fixed sets to track cost over (i.e. the elements of the cost vectors $c_k(\cdot)$). A cost vector $c_k(\cdot)$ is a partition of worlds represented by the label $\ell_k(\cdot)$ that assigns a cost to each of the disjoint sets. As we will show, the partitions are different for each vertex because we partition with respect to the new worlds that reach the given action, effect, or literal in each level. Our reason for defining the partitions this way is that the size of the partition is bounded by the number of $CLUG$ levels.

The $LUG$ and $CLUG$ construction requires first defining our initial literal layer, and then an inductive step to construct a graph level. For each graph layer of the $LUG$ and $CLUG$, we compute the label of each vertex $\ell_k(\cdot)$. In the $CLUG$, we additionally update the cost vector of the vertex. In the following we combine definitions for the $LUG$ and $CLUG$ layers, but it is easy to see that we obtain the former by omitting cost vectors and obtain the latter by computing them.

**Initial Literal Layer:** *The initial layer of the $LUG$ is defined as: $\mathcal{L}_0 = \{v_k(l) | \ell_0(l) \neq \bot\}$,
where each label is defined as: $\ell_0(l) = l \wedge BS_s$,
and each cost vector is defined as: $c_0(l) = \{\langle \ell_0(l), 0 \rangle\}$*

The $LUG$ has an initial layer, $\mathcal{L}_0$, where the label $\ell_0(l)$ of each literal $l$ represents the states of $BS_s$ in which $l$ holds. In the cost vector, we store a cost of zero for the entire group of worlds in which each literal is initially reachable (i.e. $\langle \ell_0(l), 0 \rangle$).

We illustrate the case where $BS_s = BS_I$ from the example. In Figure 2 we graphically represent the $LUG$ and index the models of $BS_s$ as worlds $\{1,2\}$. We show the cost vector $c_k(\cdot)$ for each vertex. Note, we show worlds as indexed models, but implement them using a BDD [Bryant, 1986] representation of propositional formulas. In the figure we do not explicitly show propositional labels of the elements, but do in the text. The labels for the initial literal layer are:
$\ell_0(s) = s \wedge \neg r, \ell_0(\neg s) = \neg s \wedge \neg r, \ell_0(\neg r) = \neg r$
As shown in Figure 2, the literals in the zeroth literal layer have cost zero in their initial worlds.

**Action Layer:** *The $k^{th}$ action layer of the $LUG$ is defined as: $\mathcal{A}_k = \{v_k(a) | \ell_k(a) \neq \bot\}$,
where each label is defined as: $\ell_k(a) = \ell_k^*(\rho^e(a))$,
each cost vector is defined as: $c_k(a) = \{\langle f^i(a), c^i(a) \rangle | f^i(a) \neq \bot\}$,
each cost vector partition is defined as: $f^i(a) = \ell_{k'}(a) \wedge \neg \ell_{k'-1}(a), k' \leq k$,
and each partition cost is computed as: $c^i(a) = \sum_{l \in \rho^e(a)} \text{Cover}(f^i(a), c_k(l))$*

Based on the previous literal layer $\mathcal{L}_k$, the action layer $\mathcal{A}_k$ contains all non-$\bot$ labelled causative actions from the ac-tion set $A$, plus all literal persistence. Persistence for a literal $l$, denoted by $l_p$, is represented as an action where $\rho^e(l_p) = \varepsilon^0(l_p) = l$. The label of the action at level $k$, is equivalent to the extended label of its execution precondition. We partition the cost vector based on worlds that newly support the vertex in each level. If there are new worlds supporting $a$ at level $k$, we need to add a formula-cost pair to the cost vector with the formula equal to $\ell_k(a) \wedge \neg \ell_{k-1}(a)$. When $k = 0$ we can say $\ell_{-1}(a) = \bot$. We then update the cost for each element of the cost vector. We find $c^i(a)$ by summing the costs of the execution precondition literals in the worlds described by $f^i(a)$. The cost of each literal is determined by covering the worlds $f^i(a)$ with the cost vector of the literal. In general, cost vectors do not have a specific formula-cost pair for a set of worlds we care about, rather the worlds are partitioned over several formula-cost pairs. To get a cost for the set of worlds we care about, we do a cover with the disjoint world sets in the cost vector. We try to find a minimum cost for the cover because planning graphs typically represent an optimistic projection of reachability.

**Cover(f, c):** *A Cover of a formula $f$ with a set of formula-cost pairs $c = \{\langle f^1(\cdot), c^1(\cdot)\rangle, ..., \langle f^n(\cdot), c^n(\cdot)\rangle\}$, is equivalent to a weighted set cover problem [Cormen et al., 1990] where the set of models of $f$ must be covered with weighted sets of models defined by the formula-cost pairs in c. A set of formula-cost pairs $c' \subseteq c$ covers $f$ with cost $\sum_{i:\langle f^i(\cdot), c^i(\cdot)\rangle \in c'} c^i(\cdot)$ when $f \models \bigvee_{i:\langle f^i(\cdot), c^i(\cdot)\rangle \in c'} f^i(\cdot)$.*

Finding a minimum cover is an NP-Complete problem, following from set cover. We solve it using a greedy algorithm that at each step chooses the least cost formula-cost pair that covers a new world of our set of worlds. Fortunately in the action and effect layers, the Cover operation is done with (non-overlapping) partitions, meaning there is only one possible cover. This is not the case in the literal layer construction and relaxed plan extraction because the cover is with a set of possibly overlapping sets. We show an example of using Cover after the literal layer definition.

The zeroth action layer has the following labels:
$\ell_0(B) = \ell_0(\neg r_p) = \neg r, \ell_0(C) = \ell_0(s_p) = s \wedge \neg r$,
$\ell_0(R) = \ell_0(\neg s_p) = \neg s \wedge \neg r$
The action $B$ is reachable in both worlds at a cost of zero because it has no execution precondition, whereas $C$ has a cost of zero in world two because its execution precondition holds in world two at a cost of zero.

**Effect Layer:** *The $k^{th}$ effect layer of the LUG is defined as: $\mathcal{E}_k = \{v_k(\varphi^j(a))|\ell_k(\varphi^j(a)) \neq \bot\}$,
where each label is defined as: $\ell_k(\varphi^j(a)) = \ell_k^*(\rho^j(a)) \wedge \ell_k(a)$,
each cost vector is defined as: $c_k(\varphi^j(a)) = \{\langle f^i(a), c^i(a)\rangle | f^i(a) \neq \bot\}$,
each cost vector partition is defined as: $f^i(a) = \ell_{k'}(\varphi^j(a)) \wedge \neg \ell_{k'-1}(\varphi^j(a)), k' \leq k$,
and each partition cost is compute as: $c^i(a) = c(a) + \texttt{Cover}(f^i(a), c_k(a)) + \sum_{l \in \rho^j(a)} \texttt{Cover}(f^i(a), c_k(l))$*

An effect $\varphi^j(a)$ is included in $\mathcal{E}_k$, when it is reachable in some world of $BS_s$, i.e. $\ell_k(\varphi^j(a)) \neq \bot$, which only happens when both the associated action and the antecedent are reachable in at least one world together. The cost $c^i(a)$ of world set $f^i(a)$ of an effect at level $k$ is found by adding the execution cost of the associated action, the support cost of the action in the worlds of $f^i(a)$, and the support cost of the antecedent in $f^i(a)$ (found by summing over the cost of each literal of $\varphi^j(a)$ in $f^i(a)$).

The zeroth effect layer for our example has the labels:
$\ell_0(\varphi^0(B)) = \ell_0(\varphi^0(C)) = \ell_0(\varphi^0(s_p)) = s \wedge \neg r$,
$\ell_0(\varphi^0(R)) = \ell_0(\varphi^0(\neg s_p)) = \neg s \wedge \neg r, \ell_0(\varphi^0(\neg r_p)) = \neg r$
The effect of action $B$ has the cost of $B$ in world two, even though $B$ could be executed in both worlds. This is because the effect is only enabled in world 2 by its antecedent $s$. Likewise, the effects of $C$ and $R$ have the cost of respectively executing $C$ and $R$. While not shown, the persistence effects have cost zero in the worlds of the previous level.

**Literal Layer:** *The $k^{th}$ literal layer of the LUG is defined as: $\mathcal{L}_k = \{v_k(l)|\ell_k(l) \neq \bot\}$,
where the label of each literal is defined as: $\ell_k(l) = \bigvee_{\varphi^j(a):l \in \varepsilon^j(a), v_{k-1}(\varphi^j(a)) \in \mathcal{E}_{k-1}} \ell_{k-1}(\varphi^j(a))$,
each cost vector is defined as: $c_k(l) = \{\langle f^i(a), c^i(a)\rangle | f^i(a) \neq \bot\}$,
each cost vector partition is defined as: $f^i(a) = \ell_{k'}(l) \wedge \neg \ell_{k'-1}(l), k' \leq k$,
and each partition cost is computed as: $c^i(a) = \texttt{Cover}(f^i(a), \bigcup_{\varphi^j(a):l \in \varepsilon^j(a), v_{k-1}(\varphi^j(a)) \in \mathcal{E}_{k-1}} c_{k-1}(\varphi^j(a)))$*

The literal layer, $\mathcal{L}_k$, contains all literals with non-$\bot$ labels. The label of a literal, $\ell_k(l)$, depends on $\mathcal{E}_{k-1}$ and is the disjunction of the labels of each effect that causes the literal. The cost $c^i(a)$ in a set of worlds $f^i(a)$ for a literal at level $k$ is found by covering the worlds $f^i(a)$ with the union of all formula-cost pairs of effects that support the literal.

The first literal layer for our example has the labels:
$\ell_1(s) = s \wedge \neg r, \ell_1(\neg s) = \ell_1(r) = \ell_1(\neg r) = \neg r$
In our example, we want to update the formula-cost pairs of $\neg s$ at level one. There are three supporters of $\neg s$, the persistence of $\neg s$ in world 1, the effect of action $B$ in world 2 and the $C$ action's effect in world 2. The formula-cost tuples for $\neg s$ at level 1 are for $\{1\}$ and $\{2\}$. We group the worlds this way because $\neg s$ was originally reachable in world 1, but is newly supported by world 2. For the formula-cost pair with world 1 we use the persistence in the cover. For the formula-cost pair with world 2, the supporters are $B$ and $C$, and we choose one for the cover. In Figure 2 we assign a cost of $\min(B, C)$ because we discuss two different cost models. We must also assign a cost to the formula-cost pair for $r$ in worlds 1 and 2. The cover for $r$ in these worlds must use both the effect of $C$ and $R$ because each covers only one world, hence its cost is $C + R$.

**Level off:** *The graph levels off when $\mathcal{L}_k = \mathcal{L}_{k-1}$.*

In our example we level off (terminate construction) at level two with the $LUG$ and level three with the $CLUG$. We show up to level two because level three is identical. We can say that the goal is reachable after one step from the initial belief because $BS_I = \neg r \models \ell_1^*(BS_G) = \neg r$.

## 3.2 Relaxed Plans

The relaxed plan heuristic we extract from the $LUG$ and the $CLUG$ is similar to the multiple graph relaxed plan heuristic, [Bryce and Kambhampati, 2004]. As previously described, the multiple graph heuristic uses a planning graph for every possible world of our source belief state to extract a relaxed plan to achieve the goal belief state.

The $LUG$ and $CLUG$ relaxed plan heuristics are similar by accounting for positive world interaction and independence across source states in achieving the goals. The advantage is that we find the relaxed plans by using only one planning graph to extract a single, albeit more complicated, relaxed plan. In a relaxed plan we find a line of causal support for the goal from every state in $BS_s$. Since many possible worlds use some of the same vertices to support the goal, we label relaxed plan vertices with which worlds use them. There may be several paths used to support a subgoal in the same worlds because not one is used by all worlds. For example, notice that in Figure 2 it takes both $C$ and $R$ to support r in both worlds because each action supports only one world. One challenge in extracting the relaxed plan is in tracking what worlds use which paths to support subgoals. Another challenge is in extracting cost-sensitive relaxed plans, for which the propagated cost vectors help.

The multiple graph, $LUG$, and $CLUG$ relaxed plans are inadmissible (i.e. will not guarantee optimal plans with AO* search). Admissible heuristics are lower bounds that enable search to find optimal solutions, but most in practice are very ineffective. In the next section we demonstrate that although our heuristics are inadmissible they guide our planner toward high quality solutions.

We describe relaxed plan construction by first defining relaxed plans for the $LUG$ and $CLUG$ (pointing out differences), then how the last literal layer is built, followed by the inductive step to construct a level.

**Definition 6** (Relaxed Plans). *A relaxed plan extracted from the $LUG$ or $CLUG$ for $BS_s$ is defined with respect to the goal belief state $BS_G$. The relaxed plan is a subgraph that has b levels (see below), where each level k has three layers, the literal $\mathcal{L}_k^{RP}$, action $\mathcal{A}_k^{RP}$, and effect $\mathcal{E}_k^{RP}$ layer. Each vertex $v_k^{RP}(\cdot)$ in the relaxed plan is a pair $\langle \cdot, \ell_k^{RP}(\cdot) \rangle$. The level b of the $LUG$ is the earliest level where $BS_s \models \ell_b^*(BS_G)$, and $b =$*

$$\min_k \arg\min_{S_d \in \mathcal{M}(BS_G)} \sum_{l \in S_d} \texttt{Cover}(\ell_k(l), c_k(l)),\ meaning\ every$$

*model of $BS_s$ is able to reach a model of $BS_G$ and the cost of reaching $BS_G$ is minimal.*

**Last Relaxed Plan Literal Layer:** *The final literal layer $\mathcal{L}_b^{RP}$ of the relaxed plan contains all literals that are in models of the destination belief $BS_G$. The final literal layer is a subset of the vertices in $\mathcal{L}_b$. Each literal l has a label equivalent to its label at level b, i.e. $\ell_b^{RP}(l) = \ell_b(l)$.*

**Relaxed Plan Effect Layer:** *The $k^{th}$ effect layer $\mathcal{E}_k^{RP}$ contains all the effects needed to support the literals in $\mathcal{L}_{k+1}^{RP}$. The label $\ell_k^{RP}(\varphi^j(a))$ of an effect is the disjunction of all worlds where the effect is used to support a literal. The literals in $\mathcal{L}_{k+1}^{RP}$ are supported by $\mathcal{E}_k^{RP}$ when $\forall_{v_{k+1}^{RP}(l) \in \mathcal{L}_{k+1}^{RP}}$:*

$$\ell_{k+1}^{RP}(l) \models \left( \bigvee_{\substack{\varphi^j(a): v_k^{RP}(\varphi^j(a)) \in \mathcal{E}_k^{RP}, \\ l \in \varepsilon^j(a)}} \ell_k^{RP}(\varphi^j(a)) \right)$$

The above formula states that each vertex in the literal layer must have effects chosen for the supporting effect layer such that for all worlds where the literal must be supported, there is an effect that gives support.

We construct the effect layer by using a greedy minimum cover operation for each literal to pick the effects that support worlds where the literal needs support. In the $LUG$, we use a technique that does not rely on cost vectors and at each step chooses the effect that covers the literal in the most new worlds. The intuition is that we will include less effects (and actions) if they support in more worlds. In the $CLUG$ we use a technique that at each step chooses an effect that can contribute support in new worlds at the lowest cost. We insert the chosen effects in the effect layer and label them to indicate the worlds where they were used for support.

**Relaxed Plan Action Layer:** *The $k^{th}$ action layer $\mathcal{A}_k^{RP}$ contains all actions whose effects were used in $\mathcal{E}_k^{RP}$. The associated label $\ell_k^{RP}(a)$ for each action a is the disjunction of the labels of each of its effects that are elements of $\mathcal{E}_k^{RP}$.*

**Relaxed Plan Literal Layer:** *The $k^{th}$ literal layer $\mathcal{L}_k^{RP}$ contains all literals that appear in the execution preconditions of actions in $\mathcal{A}_k^{RP}$, or the antecedents of effects in $\mathcal{E}_k^{RP}$. The associated label $\ell_k^{RP}(l)$ for each literal l is the disjunction of the labels of each of each respective action and effect in $\mathcal{A}_k^{RP}$ or $\mathcal{E}_k^{RP}$ which the literal appears in the execution precondition or antecedent.*

We support literals with effects, insert actions, and insert literals until we have supported all literals in $\mathcal{L}_1^{RP}$. Once we get a relaxed plan, the relaxed plan heuristic is the sum of the selected action costs.

In Figure 3 we show three relaxed plans to support $BS_G$

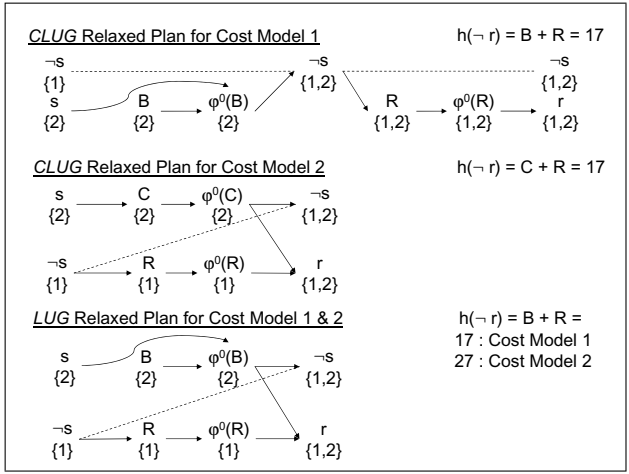

Figure 3: *Illustration of $CLUG$ and $LUG$ relaxed plans for two cost models.*

from $BS_I$. The first two are for the two cost models we presented using the $CLUG$; the third is for both cost models when using the $LUG$. All relaxed plans need to support the goal literals in worlds 1 and 2 (the worlds of $BS_I$). We find that $BS_G$ is reachable at level one, at a cost of $\min(B,C) + C + R$, and at level two at a cost of $\min(B,C)+\min(C + R,\min(B,C) + R)$. In the first cost model, level one costs 37 and level two costs 27 – so we extract starting at level 2; and with the second cost model level one costs 27 and level two costs 27 – so we extract at level 1 as there is no drop in cost at level 2. Using the $LUG$ we choose level 1 because it is the first level the goals are reachable. To extract a relaxed plan in the first cost model from the $CLUG$ we support $\neg s$ in both worlds with a persistence rather than $B$ because the persistence covers both worlds with a propagated cost of 10, opposed to 20 with $B$. Likewise, $r$ is supported with $R$ at a propagated cost of 10, opposed to 27 for the persistence. Next, we support $\neg s$ at level one in world 2 with $A$ because it is cheaper than $C$, and in world 1 with the only choice, persistence. The relaxed plan has value 17 because it chose $B$ and $R$. We leave the second cost model as an exercise. In the $LUG$ relaxed plan we could extract $B, R$ for either cost scenario because $R$ covers $r$ in the most worlds, and $B$ is chosen for supporting $\neg s$. The $LUG$ relaxed plan extraction is not sensitive to cost, but the relaxed plan value reflects action cost.

## 4 Empirical Comparisons

Our main intent is to evaluate the effectiveness of the $LUG$ and the $CLUG$ in improving the quality of plans generated by $POND$. Additionally, we also compare with two state of the art planners, GPT [Bonet and Geffner, 2000], and MBP [Bertoli *et al.*, 2001]. Even though MBP does not plan with costs, we show the cost of MBP's plans for each problem's cost model. GPT uses heuristics based on relaxing the problem to full-observability (whereas our relaxation is to no observability while ignoring action mutexes), and MBP uses a belief state's size as its heuristic merit. Our test set up involves two domains: Medical-Specialist and Rovers. Each problem had a time out of 20 minutes and a memory limit of 1GB on a 2.8GHz P4 Linux machine. We provide our planner and domain encodings at *http://rakaposhi.eas.asu.edu/belief-search/*.

POND is implemented in C and uses several existing technologies. It employs AO* search code from Eric Hansen, planning graph construction code from Joerg Hoffmann, and the BDD CUDD package from Fabio Somenzi for representing belief states, actions, and labels.

**Medical-Specialist:** We developed an extension of the medical domain [Weld *et al.*, 1998], where in addition to staining, counting of white blood cells, and medicating, one can go to a specialist for medication and there is no chance of dying – effectively allowing conformant (non-sensing) plans. We assigned costs as follows: c(stain) = 5, c(count_white_cells) = 10, c(inspect_stain) = X, c(analyze_white_cell_count) = X, c(medicate) = 5, and c(specialist_medicate) = 10. We generated ten problems, each with the respective number of diseases (1-10), in two sets where X = $\{15, 25\}$. Plans in this domain must treat a patient by either performing some combination of staining, counting white cells, and sensing actions to diagnose the exact disease and apply the proper medicate action, or using the specialist medicate action without knowing the exact disease. Plans can use hybrid strategies, using the specialist medicate for some diseases and the diagnosis and medicate for others. The strategy depends on cost and the number of diseases.

Our results in the first two columns in Figures 4, 5, and 6 show the average plan path cost, number of plan nodes (belief states) in the solution, and total time for two cost models; the x-axis reflects different problem instances. Extracting relaxed plans from the $CLUG$ instead of the $LUG$ enables $POND$ to be more cost-sensitive. The plans returned by the $CLUG$ method tend to have less nodes and a lower average path cost than the $LUG$. The $LUG$ heuristic does not measure sensing cost, but as sensing cost changes, the search is able to locally gauge the cost of sensing and adapt. Since MBP is insensitive to cost, its plans are proportionately costlier as the sensor cost increases. GPT returns better plans, but tends to take significantly more time as the cost of sensing increases; this may be attributed to how the heuristic is computed by relaxing the problem to full-observability. Our heuristics measure the cost of co-achieving the goal from a set of states, whereas GPT takes the average cost for reaching the goal from the states.

**Rovers:** We use an adaptation of the Rovers domain from the Third International Planning Competition [Long and

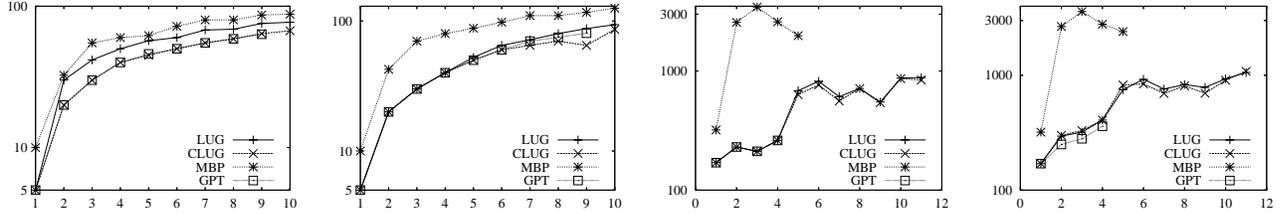

Figure 4: *Quality (average plan path cost) for $POND$ ($LUG$ and $CLUG$), $MBP$, and $GPT$ for Medical-Specialist and Rovers.*

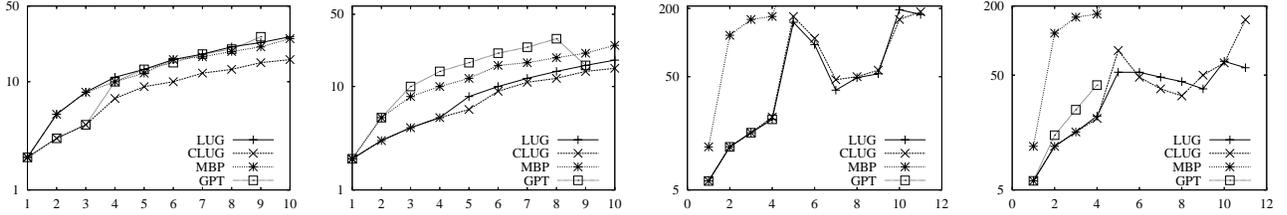

Figure 5: *Number plan nodes for $POND$ ($LUG$ and $CLUG$), $MBP$, and $GPT$ for Medical-Specialist and Rovers.*

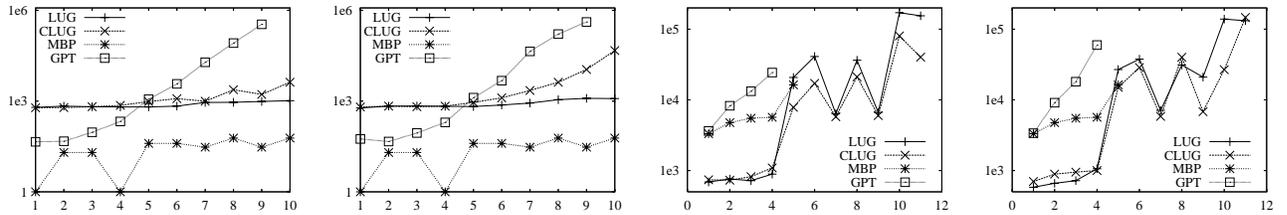

Figure 6: *Total Time(ms) for $POND$ ($LUG$ and $CLUG$), $MBP$, and $GPT$ for Medical-Specialist and Rovers.*

Fox, 2003] where there are several locations with *possible* science data (images, rocks, and soil). We added sensory actions to determine availability of scientific data and conditional actions that conformantly collect data. Our action cost model is: c(sense_visibility) = X, c(sense_rock) = Y, c(sense_soil) = Z, c(navigate) = 50, c(calibrate) = 10, c(take_image) = 20, c(communicate_data) = 40, c(sample_soil) = 30, c(sample_rock) = 60, and c(drop) = 5. The two versions have costs: (X,Y,Z) = {(35, 55, 45), (100, 120, 110)}. Plans in the rovers domain can involve sensing at locations to identify if data can be collected or simply going to every possible location and trying to collect data. The number of locations varies between four and eight, and the number of possible locations to collect up to three types of data can be between one and four.

The last two columns of Figures 4, 5, and 6 show the average path cost, number of nodes in the solution, and total time for the two cost models. We found that the $LUG$ and $CLUG$ relaxed plan extraction guide POND toward similar plans, in terms of cost and number of nodes. The lack of difference between the heuristics may be attributed to the domain structure – good solutions have a lot of positive interaction (i.e. the heuristics extract similar relaxed plans because low cost actions also support subgoals in many possible worlds), opposed to Medical where solutions are fairly independent for different possible worlds. MBP, making no use of action costs, returns plans with considerably (a order of magnitude) higher average path costs and number of solution nodes. GPT fares better than MBP in terms of plan cost, but both are limited in scalability due to weaker heuristics.

In summary, the experiments show that the $LUG$ and $CLUG$ heuristics help with scalability and that using the $CLUG$ to extract relaxed plans can help find better solutions. We also found that planners not reasoning about action cost can return arbitrarily poor solutions, and planners whose heuristic relaxes uncertainty do not scale as well.

## 5 Related Work

The idea of cost propagation on planning graphs was first presented by Do and Kambhampati [2003] to cope with metric-temporal planning. The first work on using planning graphs in conditional planning was in the CGP [Smith and Weld, 1998] and SGP [Weld *et al.*, 1998] planners. Recently, planning graph heuristics have proven useful in conformant planning [Bryce and Kambhampati, 2004; Brafman and Hoffmann, 2004] and conditional planning [Cushing and Bryce, 2005; Hoffmann and Brafman, 2005]. They have also proven useful in reachability analysis for MDPs [Boutilier *et al.*, 1998]; our work could be extended for POMDPs. Also related is the work on sensor planning, such as Koenig and Liu [1999]. The authors investigate the frequency of sensing as the plan optimization criterion changes (from minimizing the worst case cost to the expected cost). We investigate the frequency of sensing while minimizing average plan cost under different cost models. The work on optimal limited contingency planning

[Meuleau and Smith, 2003] stated that adjusting sensory action cost, as we have, is an alternative to their approach for reducing plan branches.

## 6 Conclusion & Future Work

With our motivation toward conditional planning approaches that can scale like classical planners, but still reason with quality metrics like POMDPs, we have presented a novel planning graph generalization called the $LUG$ and an associated cost propagated version called the $CLUG$. With the $CLUG$ we extract cost-sensitive relaxed plans that are effective in guiding our planner $POND$ toward high-quality conditional plans. We have shown with an empirical comparison that our approach improves the quality of conditional plans over conditional planners that do not account for cost information, and we that can out-scale approaches that consider cost information and uncertainty in a weaker fashion.

While our relaxation of conditional planning ignores sensory actions, we have explored techniques to include observations in heuristic estimates. The basic idea is to extract a relaxed plan then add sensory actions that reduce cost by removing mutexes (sensing to place conflicting actions in different branches) or reducing average path cost (ensuring costly actions are not executed in all paths). The major reason we do not report on using sensory relaxed plans here is that scalability of these techniques is somewhat limited, despite their ability to further improving plan quality. We are investigating ways to reduce computation cost.

Given our ability to propagate numeric information on the $LUG$, we are currently adapting these heuristics and our planner to handle non uniform probabilities. The extension involves adding probabilities to labels by using ADDs instead of BDDs, and redefining propagation semantics. The propagation semantics replaces conjunctions with products, and disjunctions with summations. A label represents a probability distribution over possible worlds, the probability of reaching the vertex is a summation over the possible world probabilities, and the expected cost of a vertex is the sum of products between cost vector partitions and the label. Relaxed plans, which previously involved weighted set covers with a single objective (minimizing cost) become multi-objective by trading off cost and probability.

In addition to cost propagation we have also extended the $LUG$ within the framework of state agnostic planning graphs [Cushing and Bryce, 2005]. The $LUG$ seeks to avoid redundancy across the multiple planning graphs built for states in the same belief state. We extended this notion to avoid redundancy in planning graphs built for every belief state. We have shown that the state agnostic $LUG$ ($SLUG$) which is built once per search episode (opposed to a $LUG$ at each node) can reduce heuristic computation cost without sacrificing informedness.

**Acknowledgements:** This research is supported in part by the NSF grant IIS-0308139 and an IBM Faculty Award to Subbarao Kambhampati.